\documentclass[sigconf, nonacm]{acmart}

\AtBeginDocument{%
  }

\acmConference[RelKD '26]{International Workshop on Resource-Efficient Learning for Knowledge Discovery}{August 2026}{Jeju, South Korea}

\setcopyright{none}
\makeatletter
\let\@isbn\@empty
\let\@doi\@empty
\makeatother

\begin{document}

\title{MORL-A2C: Multi-Objective Reinforcement Learning Reranker for
  Optimizing Healthiness in MOPI-HFRS}

\author{Aarya Vasantlal}
\affiliation{%
  \institution{University of Connecticut}
  \city{Storrs}
  \state{CT}
  \country{USA}}
\email{aarya.vasantlal@uconn.edu}

\author{Joshua Zolla}
\affiliation{%
  \institution{University of Connecticut}
  \city{Storrs}
  \state{CT}
  \country{USA}}
\email{joshua.zolla@uconn.edu}

\author{Chuxu Zhang}
\affiliation{%
  \institution{University of Connecticut}
  \city{Storrs}
  \state{CT}
  \country{USA}}
\email{chuxu.zhang@uconn.edu}

\renewcommand{\shortauthors}{Vasantlal, Zolla, and Zhang}

\begin{abstract}
Unhealthy dietary behavior continues to be a persistent and escalating public health issue in the United States, exacerbated by the prevalence of recommendation systems that prioritize user preference without adequately considering nutritional health. The Multi-Objective Personalized Interpretable Health-aware Food Recommendation System (MOPI-HFRS) \cite{MOPI_HFRS}, from which this work extends, addresses this limitation by jointly optimizing user preference, personalized health, and nutritional diversity through a Pareto-based optimization framework. However, this approach relies on static, per-step tradeoff solutions which fail to capture the sequential and cumulative nature of dietary decision-making, in which the value of recommending an item depends on what has already been recommended. In this work, we introduce the Multi-Objective Reinforcement Learning Reranker (MORL-A2C), a sequential decision-making extension to MOPI-HFRS that targets the health-preference axis of the multi-objective problem. Leveraging frozen graph neural network embeddings from MOPI-HFRS, MORL-A2C formulates recommendation as a $K$-step sequential reranking problem, employing an Advantage Actor-Critic (A2C) algorithm to learn a policy balancing relevance and health objectives through a scalarized reward signal. To accelerate convergence and avoid degenerate exploration, the policy is warm-started using behavior cloning pretraining against a strong dot-product ranker derived from the frozen embeddings. In the course of this work, we identify and correct a non-trivial bug in the original MOPI-HFRS evaluation pipeline that understated baseline ranking performance; all RL-based metrics are reported against the corrected baseline. Experimental results on both the macro-nutrient and full-nutrient benchmarks indicate that MORL-A2C achieves a modest reduction in ranking quality relative to the corrected baseline (Recall@20: $25.64\% \rightarrow 23.61\%$, NDCG@20: $23.52\% \rightarrow 20.64\%$ on the macro benchmark) in exchange for a substantial improvement in personalized health alignment (H-Score@20: $46.05\% \rightarrow 69.57\%$), with consistent trends on the full-nutrient benchmark. These findings validate that policy-driven sequential optimization can effectively navigate the health-preference trade-off in multi-objective food recommendation. Our code is publicly available at \url{https://github.com/JoshZ411/MOPI-HFRS-SLC}.
\end{abstract}




\settopmatter{printacmref=false} 
\makeatletter
\gdef\@copyrightbox{}            
\makeatother

\maketitle

\section{Introduction}

Diet-related health challenges, while often preventable through healthy eating habits, continue to intensify in the United States. CDC reports indicate that in 2023, every U.S.\ state had an adult obesity prevalence higher than 20\%, with 23 states above 35\% and three states above 40\%\cite{cdc_obesity}. Compounding this is a widespread failure to meet baseline nutritional recommendations: in 2019, only 12.3\% of adults met fruit-intake recommendations and only 10.0\% met vegetable recommendations\cite{cdc_fruitveg}. As a consequence, 42\% of Americans live with at least two chronic conditions and 12\% with at least five~\cite{cdc_chronic}. Despite increasing awareness, existing food recommendation platforms still optimize predominantly for engagement and short-term user preference, neglecting the longer-term health consequences of dietary choices.

The Multi-Objective Personalized Interpretable Health-aware Food Recommendation System (MOPI-HFRS), the primary system upon which this work extends, addresses many of these challenges~\cite{MOPI_HFRS}. First, it provides an adaptive, per-user interpretation of food healthiness influenced by prior dietary habits, clinical metrics, and biometric data, leveraging the National Health and Nutrition Examination Survey (NHANES). Second, it integrates a downstream LLM-enhanced reasoning module to provide human-interpretable explanations of recommendations. Third, and most relevant to this work, it intelligently balances three competing dietary considerations (preference, personalized health, and nutritional diversity) during recommendation. To do so, MOPI-HFRS employs Pareto-based multi-objective optimization via the Multiple-Gradient Descent Algorithm (MGDA)~\cite{MOPI_HFRS}, computing a non-dominated descent direction at each training step.

Although effective in producing locally non-dominated solutions, this approach is fundamentally myopic; it treats each optimization step as an independent problem and does not account for the sequential nature of recommendation decisions. The value of recommending a particular food at position $t$ depends on what has already been recommended at positions $1, \dots, t-1$, a cumulative coverage signal that static one-shot optimization cannot capture. Additionally, MGDA optimizes losses, not the actual ranking metrics or health alignment metrics ultimately reported at evaluation time. To address these limitations, we propose the Multi-Objective Reinforcement Learning Reranker (MORL-A2C), a novel extension that reframes the health-preference component of the recommendation problem as a sequential decision-making task. Rather than selecting items via static optimization, MORL-A2C learns a policy that constructs a ranked list of $K$ recommendations over multiple steps, optimizing for long-term cumulative reward along the relevance and health objectives. This enables the modeling of interactions between items within a recommendation list and the capture of delayed tradeoffs between user preference and personalized health alignment.

Our framework builds on the existing MOPI-HFRS architecture by treating its pretrained user and food GNN embeddings as a fixed representation space. On top of this representation, we introduce a reinforcement learning agent trained using an Advantage Actor-Critic (A2C) algorithm. The agent interacts with a deterministic, simulated environment that evaluates recommendations against held-out positives and user health tags, producing a scalarized two-component reward signal that guides policy learning. The policy is warm-started via behavior cloning from a strong dot-product ranker, allowing the system to begin training from a high-quality initialization rather than exploring randomly over the very large food action space. We note that the current formulation focuses on the health-preference trade-off; incorporating diversity as an explicit reward component is a natural extension discussed in Section~7.

Notably, in the course of evaluating MORL-A2C, we discovered and corrected a non-trivial bug in the original MOPI-HFRS evaluation pipeline that affected reported ranking metrics. We report all results against this corrected baseline, and the experimental section explicitly compares (i) the original MOPI-HFRS reported metrics, (ii) the bug-fixed MOPI-HFRS metrics, and (iii) MORL-A2C metrics, to provide an honest and fair picture of where the proposed system stands.

\section{Related Works}

\subsection{Multi-Objective Recommender Systems}
Modern recommender systems, including the food recommendation problem we address, frequently sit on the active research front of Graph Neural Network (GNN)-based recommendation. As in MOPI-HFRS, user--item interactions are represented as edges in a bipartite graph, and refined embeddings are produced by propagating relational information through the graph structure. Simpler systems optimize a single objective such as top-$K$ relevance ranking, but many real-world deployments require balancing multiple objectives that may be in tension. Such problems are referred to as Multi-Objective Recommender Systems (MORS) problems~\cite{MORS_Challenges}. Approaches range from linear scalarization of multiple losses, to constrained optimization, to Pareto-based methods that aim for non-dominated solutions in objective space. Each has tradeoffs in terms of weight sensitivity, computational cost, and ability to express objective interactions.

\subsection{MOPI-HFRS Pareto Optimization}
MOPI-HFRS applies Pareto optimization via MGDA, which computes a Pareto-feasible descent direction at each training step by minimizing the norm of a convex combination of objective gradients. Three losses are jointly optimized.

To encourage user preference, the framework defines a Bayesian Personalized Ranking (BPR) loss:
\[
\mathcal{L}_{\mathrm{BPR}}
= - \sum_{(u,i,j)} \ln \sigma(\hat{y}_{uij})
+ \lambda \left\| \Theta \right\|^{2},
\]
where $\hat{y}_{uij} = \hat{y}_{ui} - \hat{y}_{uj}$ is the predicted score difference between a positively interacted food $i$ and a sampled non-interacted food $j$ for user $u$.

Personalized health is captured through tag overlap:
\[
\mathcal{L}_{\mathrm{health}}
= - \sum_{(u,i,j)}
    \ln \big(
        ( \mathcal{I}(t_u, t_i) - \mathcal{I}(t_u, t_j) )
        \cdot \sigma(\hat{y}_{uij})
    \big),
\]
where $\mathcal{I}(t_u, t_\cdot)$ denotes Jaccard similarity between user and food health-tag vectors. 

Diversity is captured by a third loss penalizing pairwise embedding similarity within the recommended set:
\[
\mathcal{L}_{\mathrm{diversity}}
= - \sum_{(u,i,j)}
    \ln \sigma \Bigg(
        1 - \sum_{i,j \in \mathcal{N}_u}
                \cos(\mathbf{f}_i, \mathbf{f}_j) \cdot \hat{y}_{ui}
    \Bigg),
\]
where \(\mathcal{N}_u\) is the set of foods recommended to user \(u\), \(\mathbf{f}_i\) and \(\mathbf{f}_j\) are their embedding vectors.
MGDA solves
\[
\mathcal{L}_{\mathrm{Pareto}}
    = \min_{\{\alpha_k\}}
        \left\lVert
            \sum_{k=1}^{K}
                \alpha_k \, \nabla_{\theta} \mathcal{L}_k(\mathcal{G}; \theta)
        \right\rVert
\]
to find the minimum-norm descent direction inside the convex hull of objective gradients. While this guarantees non-dominated solutions \emph{locally}, it does not account for sequential dependencies across positions in a recommendation list, nor does it directly optimize ranking-quality metrics that are inherently non-differentiable. Of these three objectives, the health-preference trade-off ($\mathcal{L}_{\mathrm{BPR}}$ and $\mathcal{L}_{\mathrm{health}}$) is the focus of our RL extension; diversity optimization via $\mathcal{L}_{\mathrm{diversity}}$ is left to future work (Section~7).

\subsection{Reinforcement Learning for Recommendation}
Reinforcement Learning (RL) is a framework in which an agent learns an optimal policy by interacting with an environment. At each step, the agent observes a state, selects an action via an exploration policy, and receives a reward; over time, it adjusts a policy $\pi_\theta$ to maximize expected cumulative discounted reward $\mathbb{E}\!\left[\sum_{t=0}^{T} \gamma^{t} r_t\right]$. Unlike traditional supervised optimization, RL methods can model delayed and compounding effects, making them well-suited to recommendation tasks where actions interact: session-based recommendation, slate optimization, and sequential item selection have all benefited from RL formulations~\cite{RLforRecSys}.

\subsection{Multi-Objective Reinforcement Learning}
Extending RL to multiple objectives introduces vector-valued rewards $\mathbf{v}_t = (v_{t,1}, \dots, v_{t,K})$. Multi-Objective Reinforcement Learning (MORL) addresses optimization in this setting, often through scalarization functions that aggregate components into a single scalar reward, parameterized by a weight vector $\mathbf{w}$:
\[
v_{\mathbf{w}} = f(\mathbf{v}, \mathbf{w}) = \sum_{k=1}^{K} w_{k} v_{k}.
\]
Multi-policy MORL methods aim to approximate the entire Pareto frontier of optimal policies~\cite{MORL_Pareto}, but are typically computationally infeasible at the scale of large catalogs and large embedding spaces. As such, single-policy MORL with scalarized rewards is the predominant practical alternative, with extensions including dynamic and state-dependent weighting~\cite{DRS}. We adopt a single-policy, fixed-coefficient scalarization in this work as a tractable starting point.

\subsection{Policy Gradient Methods and Actor-Critic Learning}
Policy gradient methods directly optimize a parameterized policy $\pi_\theta(a \mid s)$ via gradient ascent on expected return. The classical REINFORCE algorithm uses
\[
\nabla J(\theta) = \mathbb{E}\!\left[ G_t \cdot \nabla_\theta \log \pi_\theta(a_t \mid s_t) \right],
\]
where $G_t = \sum_{t'=t}^{T} \gamma^{t'} r_{t'}$ is the cumulative discounted return. Although conceptually simple, REINFORCE suffers from high variance, particularly when reward magnitudes vary substantially or rewards are sparse, both of which are characteristic of multi-objective recommendation, where scalarized reward magnitudes shift with weight choices.

Actor-critic methods extend policy gradients by introducing a learned value function $V_\phi(s)$ that serves as a state-dependent baseline, replacing the raw return with the advantage
\[
A_t = G_t - V_\phi(s_t),
\]
yielding the lower-variance gradient estimator
\[
\nabla J(\theta) = \mathbb{E}\!\left[ A_t \cdot \nabla_\theta \log \pi_\theta(a_t \mid s_t) \right].
\]
The Advantage Actor-Critic (A2C) framework jointly trains an actor (policy) and a critic (value head) by minimizing

\[
\mathcal{L}
=
\underbrace{-\sum_t A_t \log \pi_\theta(a_t \mid s_t)}_{\mathcal{L}_{\text{policy}}}
+
\lambda_v \underbrace{\,\mathrm{MSE}(V_\phi(s_t),\, G_t)}_{\mathcal{L}_{\text{value}}}
-
\lambda_e \underbrace{\,\mathcal{H}(\pi_\theta(\cdot \mid s_t))\,}_{\text{entropy}},
\]
where $\lambda_v$ and $\lambda_e$ control the relative weight of value estimation and entropy-based exploration. A2C is particularly well-suited to recommendation problems with very large discrete action spaces, where it avoids the need for explicit per-step maximization over all actions, as required in value-based methods such as DQN.

\subsection{Behavior Cloning Pretraining}
A persistent challenge in applying RL to large-catalog recommendation is that policies trained from random initialization rarely escape early-training collapse. The action space spans the entire food catalog, the base reward is sparse, and entropy-driven exploration alone is insufficient to consistently discover the small subset of items worth ranking highly for any given user. The standard remedy is behavior cloning (BC): a supervised pretraining stage in which the policy is trained to imitate the decisions of a strong reference model before any reinforcement signal is applied~\cite{PIRLNav}.

In recommendation specifically, a pretrained ranking model is a natural source of supervision: it encodes which items are plausible candidates for each user, and its top-ranked predictions provide dense per-step targets that the eventual RL policy can refine rather than rediscover. This warm-start is particularly important in multi-objective settings, where unguided exploration tends to converge on degenerate policies that maximize only one reward component; for instance, repeatedly selecting items with broad health-tag coverage regardless of whether the user would ever interact with them. By grounding the policy near a strong relevance prior, BC pretraining ensures that subsequent multi-objective RL refinement perturbs around a sensible baseline rather than searching from scratch in an essentially unstructured action space.

\section{Problem Formulation}

We now formalize the recommendation problem addressed by MORL-A2C. Our goal is to leverage the strengths of MOPI-HFRS while replacing its static GNN-based ranking with a sequential RL policy that jointly optimizes relevance and health alignment through a scalarized per-step reward, using frozen GNN embeddings as input.

\subsection{Graph and Representation Preliminaries}

Let $\mathcal{U}$ denote the set of users and $\mathcal{F}$ denote the set of food items. Interaction data is represented as the bipartite graph
\[
\mathcal{G} = (\mathcal{U}, \mathcal{F}, \mathcal{E}, X_{\mathcal{U}}, X_{\mathcal{F}}),
\]
where $(u, f) \in \mathcal{E}$ indicates a user--food interaction and $X_{\mathcal{U}}, X_{\mathcal{F}}$ contain user demographic / health features and food nutritional attributes respectively. A graph neural encoder $F_{\Theta}$, identical to the SGSL-based encoder in MOPI-HFRS, produces user and food embeddings
\[
E^{\text{user}}_{\mathrm{final}} \in \mathbb{R}^{|\mathcal{U}| \times d}, \qquad
E^{\text{food}}_{\mathrm{final}} \in \mathbb{R}^{|\mathcal{F}| \times d},
\]
which serve as the representational backbone of the RL environment. In MORL-A2C, $F_\Theta$ is trained as in MOPI-HFRS and then \emph{frozen}; the RL agent operates over a fixed embedding space, focusing all learning capacity on policy optimization rather than representation learning.

\subsection{Sequential Recommendation as a Markov Decision Process}

\begin{figure}[htbp]
    \centering
    \includegraphics[width=0.95\columnwidth]{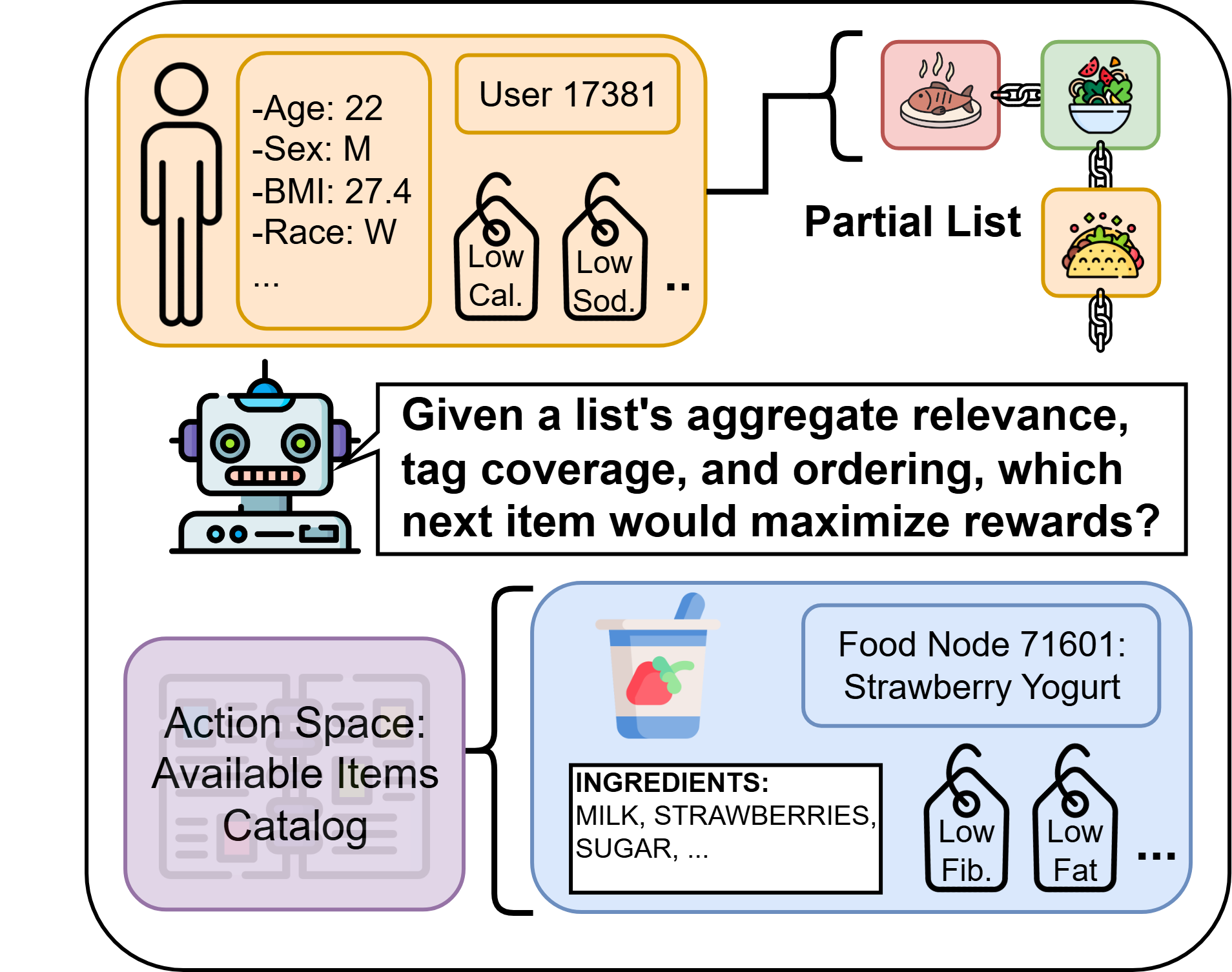}
    \caption{Conceptual illustration of the MORL-A2C agent's decision setting. For each user (top: ID, demographics, clinical info, and health tags), the agent selects from an action space comprising the available food catalog. Each food item carries descriptive features (ingredients, category) and a set of nutritional health tags. At each step, the agent chooses one item to add to the user's top-$K$ recommendation list, with the choice informed by the frozen GNN embeddings of the user and the candidate items.}
    \Description{An illustration of the agent's decision setting, showing a user profile with demographic and health tag information on the top left, a robot icon representing the agent in the middle, an action space of available food items on the bottom left, and an example food node (strawberry yogurt) with ingredients and nutritional health tags on the bottom right.}
    \label{fig:agent-formulation}
\end{figure}

We reformulate the multi-objective recommendation task as a Markov Decision Process (MDP)
\[
(\mathcal{S}, \mathcal{A}, P, \mathcal{R}, \gamma),
\]
in which the agent constructs the top-$K$ recommendation list for each user one item at a time. This stands in contrast to MOPI-HFRS, where the entire top-$K$ list is produced in a single, embedding-based scoring pass. Figure~\ref{fig:agent-formulation} illustrates this setup at a high level: the agent selects items from the food catalog one at a time on behalf of each user, conditioned on user demographics, health tags, and item nutritional attributes.

\paragraph{State Space $\mathcal{S}$.}
At step $t \in \{0, 1, \dots, K-1\}$ within a user's episode, the state $s_t \in \mathcal{S}$ encodes both the user and the partially-constructed recommendation list:
\[
s_t = \big[\, h_u \,;\; \bar{e}_t \,;\; \mathbf{c}_t \,;\; t/K \,\big] \in \mathbb{R}^{2d + |T| + 1},
\]
where:
\begin{itemize}
    \item $h_u \in \mathbb{R}^d$ is the frozen user embedding from $E^{\text{user}}_{\mathrm{final}}$;
    \item $\bar{e}_t \in \mathbb{R}^d$ is the running mean of the embeddings of items selected at positions $1, \dots, t$;
    \item $\mathbf{c}_t \in \{0,1\}^{|T|}$ is the cumulative \emph{tag coverage} vector, recording which health tags have been touched by any item already selected;
    \item $t/K$ is a normalized timestep encoding.
\end{itemize}
This state representation explicitly encodes the structure of sequential recommendation: which items are still ``remaining'' (via $\bar{e}_t$ and $\mathbf{c}_t$), how far along the agent is in the list, and what the user looks like in embedding space. Unlike a contextual bandit, the state evolves nontrivially across the episode, allowing the policy to reason about \emph{position-dependent} value.

\paragraph{Action Space $\mathcal{A}$.}
At each step $t$, the agent selects one food item to add to the list. Items already selected during the current episode are removed from the candidate set, so $\mathcal{A}_t = \mathcal{F} \setminus \{a_0, \dots, a_{t-1}\}$. The default candidate pool is the full food catalog.

\paragraph{Transition Function $P$.}
Transitions are deterministic: given $s_t$ and selected action $a_t$,
\[
\bar{e}_{t+1} = \frac{t \cdot \bar{e}_t + e_{a_t}}{t+1}, \qquad
\mathbf{c}_{t+1} = \mathrm{clip}\!\left(\mathbf{c}_t + \mathbf{t}_{a_t},\, 0,\, 1\right),
\]
and the candidate set is updated by removing $a_t$. Episodes terminate after $K$ items have been selected.

\paragraph{Reward Function $\mathcal{R}$.}
At each step, the environment returns a two-component reward:
\[
\mathbf{r}_t = (r^{\mathrm{rel}}_t,\, r^{\mathrm{health}}_t),
\]
where $r^{\mathrm{rel}}_t = \mathbf{1}[a_t \in \mathcal{P}_u^{\text{train}}]$ rewards items in the user's training-time positive set, and $r^{\mathrm{health}}_t = \mathbf{1}[\mathbf{t}_{a_t} \wedge \mathbf{t}_u \neq \mathbf{0}]$ rewards items sharing at least one health tag with the user. The two components are scalarized via a fixed coefficient $\beta$:
\[
r_t = r^{\mathrm{rel}}_t + \beta \cdot r^{\mathrm{health}}_t.
\]
The relative weighting controls the relevance--health tradeoff and is the primary lever for shaping policy behavior.

\paragraph{Discount Factor $\gamma$.}
Future rewards are discounted by $\gamma \in (0, 1]$, reflecting that recommendations earlier in the list contribute most directly to evaluation metrics such as NDCG@$K$.

\subsection{Learning Objective}

The agent learns a stochastic policy $\pi_\theta(a \mid s)$ that maximizes the expected discounted return
\[
J(\theta) = \mathbb{E}_{u \sim \mathcal{U},\, \tau \sim \pi_\theta}\!\left[\sum_{t=0}^{K-1} \gamma^t r_t\right],
\]
where $\tau = (s_0, a_0, \dots, s_{K-1}, a_{K-1})$ is a trajectory generated by rolling out the policy from a sampled user. Because the agent makes $K$ decisions per user with each decision conditioned on previous selections, the policy can shape \emph{set-level} properties of the recommendation list (tag coverage, item diversity, position-aware health balance, etc.) that one-shot rankers cannot natively express.

\subsection{Why Sequential Optimization Matters}

Static, embedding-based scoring of the form $\hat{y}_{ui} = \langle h_u, e_i \rangle$, as used in MOPI-HFRS, makes per-item decisions that are independent across positions. Once embeddings are fixed, the top-$K$ list is a deterministic argsort. Two consequences follow:

\begin{enumerate}
    \item \textbf{No interaction between recommendations.} If the top scoring item already covers a critical health tag, the second item still receives no value for covering that tag, even though doing so adds nothing on the margin. Conversely, an item that would fill an uncovered tag gap receives no extra credit.
    \item \textbf{No optimization of non-differentiable metrics.} Ranking metrics (Recall@$K$, NDCG@$K$) and tag-overlap-style health metrics (H-Score@$K$) are not directly differentiable. Static optimization minimizes \emph{surrogate} losses (BPR, Jaccard-weighted BPR, cosine-similarity penalty) and relies on these surrogates to correlate with the actual evaluation criteria.
\end{enumerate}

By formulating recommendation as a $K$-step MDP, MORL-A2C optimizes \emph{policies}, or mappings from states to action distributions, against a reward signal that more directly reflects the evaluation criteria of the system. The state explicitly carries the partially-constructed list, so the policy can adapt its choice at position $t$ based on what is already covered.

\section{Methodology and Model}

This section describes the architecture and training procedure for MORL-A2C: (i) a fixed graph encoder inherited from MOPI-HFRS that produces user and food embeddings; (ii) a sequential environment exposing the MDP defined above; and (iii) an actor-critic learner with behavior cloning pretraining that produces the deployment-time policy.

\begin{figure*}[t]
    \centering
    \includegraphics[width=0.7\textwidth]{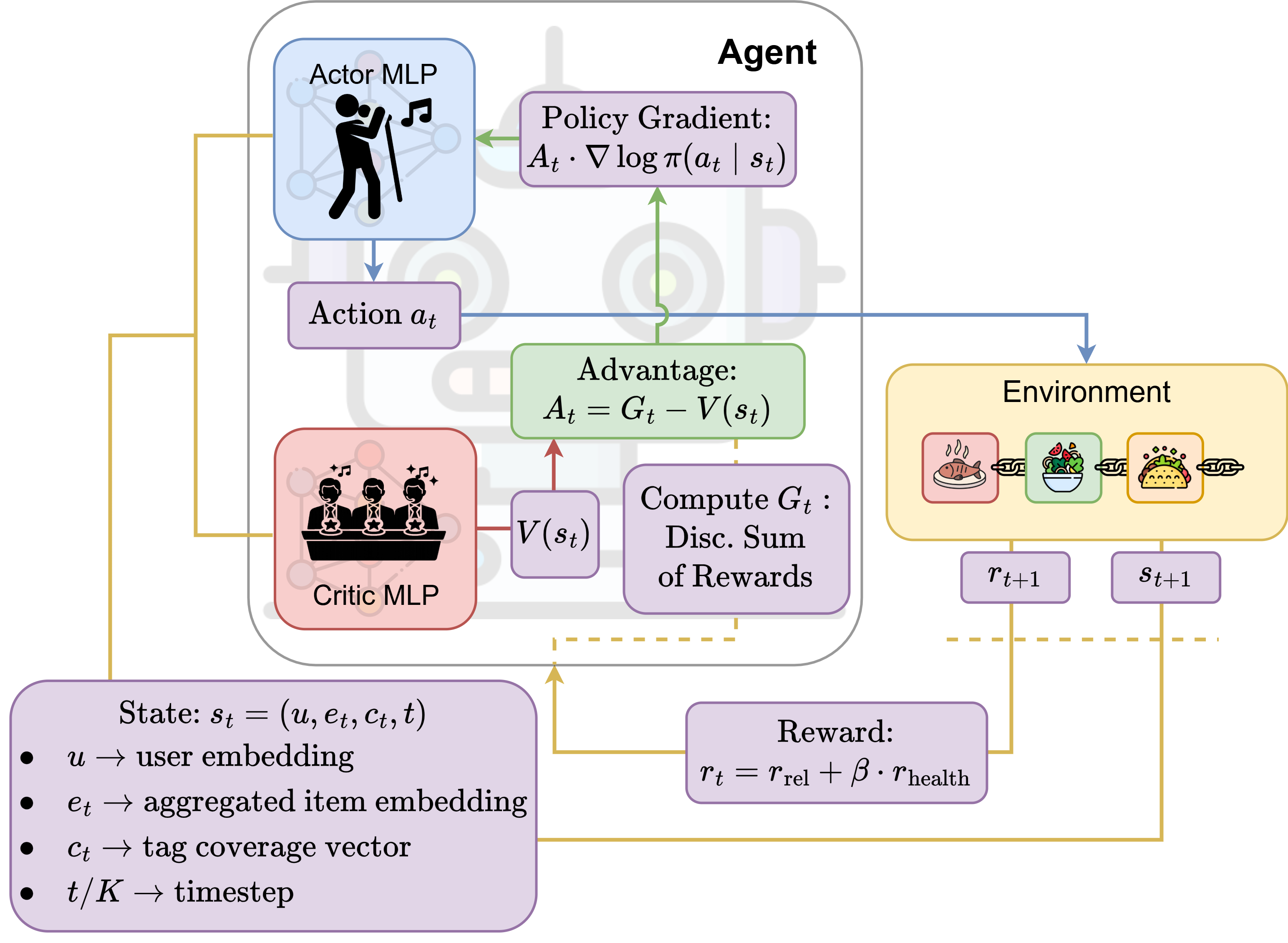}
    \caption{Advantage Actor-Critic loop for MORL-A2C. The actor MLP samples action $a_t$ from $\pi_\theta(\cdot \mid s_t)$; the environment returns the scalarized reward $r_t = r_{\mathrm{rel}} + \beta \cdot r_{\mathrm{health}}$ and the next state $s_{t+1}$, where the state components $u, e_t, c_t, t$ correspond to $h_u, \bar{e}_t, \mathbf{c}_t, t/K$ respectively in the notation of Section~3.2. Discounted returns $G_t$ are combined with the critic's value estimate $V_\phi(s_t)$ to produce the advantage $A_t$, which scales the policy gradient. Actor weights are initialized via the behavior cloning stage of Section~\ref{subsec:bc-pretraining} so that A2C refinement begins from a policy that already replicates the dot-product reference ranker.}
    \Description{A diagram of the Advantage Actor-Critic training loop, showing the actor MLP sampling an action, the environment returning a scalarized reward and next state, and the critic producing a value estimate used to compute the advantage that scales the policy gradient.}
    \label{fig:a2c-loop}
\end{figure*}

\subsection{Frozen GNN Encoder}

The base GNN encoder $F_\Theta$ is trained exactly as in MOPI-HFRS, using its joint preference / health / diversity loss formulation, and then frozen. A forward pass yields $E^{\text{user}}_{\mathrm{final}}$ and $E^{\text{food}}_{\mathrm{final}}$, which the agent treats as fixed feature tensors. This decouples representation learning from policy learning, both lowering training cost and providing a stable embedding geometry against which the policy can specialize. Concretely, the actor and critic MLPs together comprise approximately 271k trainable parameters, while the frozen GNN encoder contains nearly 3.9M; MORL-A2C thus trains fewer than 7\% of the parameters in the base model, making the approach well-suited to resource-constrained settings where the full recommender is expensive to retrain.

\subsection{Actor and Critic Architectures}

The policy network (actor) takes the current state $s_t$ and a candidate matrix $C_t \in \mathbb{R}^{|\mathcal{A}_t| \times d}$ representing the embeddings of remaining items, and produces logits over the candidate set. Concretely:
\begin{align*}
\mathbf{h}_s &= \mathrm{MLP}_{\mathrm{state}}(s_t),\\
\mathbf{H}_c &= \mathrm{MLP}_{\mathrm{cand}}(C_t),\\
\ell_t &= \mathbf{H}_c \, \mathbf{h}_s, \qquad \pi_\theta(\cdot \mid s_t) = \mathrm{softmax}(\ell_t),
\end{align*}
where $\mathrm{MLP}_{\mathrm{state}}$ is a two-layer ReLU network mapping the $(2d + |T| + 1)$-dimensional state into a hidden space and $\mathrm{MLP}_{\mathrm{cand}}$ is a single-layer ReLU encoder applied per candidate. Logits are computed as inner products between the encoded state and each encoded candidate. During training, actions are sampled from $\pi_\theta$; at evaluation time, actions are taken greedily.

The critic is a separate MLP $V_\phi : \mathbb{R}^{2d + |T| + 1} \to \mathbb{R}$ producing scalar state-value estimates. It receives only the state, not the candidate set, since under the policy distribution, candidate-set effects are already integrated into the expected return.

\subsection{Reward Scalarization}

The two-component per-step reward is scalarized using a fixed coefficient:
\[
r_t = r^{\mathrm{rel}}_t + \beta \cdot r^{\mathrm{health}}_t.
\]
Both $r^{\mathrm{rel}}_t$ and $r^{\mathrm{health}}_t$ lie in $\{0, 1\}$, so $r_t \in \{0, 1, \beta, 1 + \beta\}$. Increasing $\beta$ shifts policy preference toward items that match the user's health-tag profile, even at the cost of some relevance. We treat $\beta$ as a designer-selected hyperparameter; the question of learning a state-dependent $\beta(s_t)$, as in dynamic reward scalarization~\cite{DRS}, is left to future work.

\subsection{A2C Training}

For each training epoch, MORL-A2C samples a batch of users and rolls out one $K$-step episode per user, following $\pi_\theta$. Importantly, $\pi_\theta$ enters this stage already initialized via behavior cloning (Section~\ref{subsec:bc-pretraining}), so the rollouts driving A2C updates from epoch 1 already reflect strong relevance behavior; A2C refinement perturbs this baseline rather than discovering it. From each trajectory, discounted returns
\[
G_t = \sum_{t'=t}^{K-1} \gamma^{\,t'-t} r_{t'}
\]
are computed, and per-step advantages are formed using the critic:
\[
A_t = G_t - V_\phi(s_t).
\]
The policy and value heads are updated jointly by minimizing the standard A2C objective:
\[
\begin{aligned}
\mathcal{L}(\theta, \phi) =
&\underbrace{- \sum_t A_t \log \pi_\theta(a_t \mid s_t)}_{\mathcal{L}_{\mathrm{policy}}} \\
&+ \lambda_v \underbrace{\sum_t (V_\phi(s_t) - G_t)^2}_{\mathcal{L}_{\mathrm{value}}} \\
&- \lambda_e \underbrace{\sum_t \mathcal{H}(\pi_\theta(\cdot \mid s_t))}_{\text{entropy}}.
\end{aligned}
\]
The entropy term encourages exploration over the candidate set, particularly important early in training when the policy might otherwise collapse onto a small number of high-relevance items and never discover health-aligned alternatives. Gradients are clipped and applied via a standard first-order optimizer; the GNN encoder is \emph{not} updated during this phase.

\subsection{Behavior Cloning Pretraining}
\label{subsec:bc-pretraining}

To avoid the failure modes described in Section~2.6, we precede A2C training with a behavior cloning stage in which the policy network is trained to reproduce the recommendations of a strong reference ranker derived from the frozen MOPI-HFRS embeddings. The reference ranker scores each user--item pair by inner product, $\hat{y}_{ui} = \langle h_u, e_i \rangle$, and the GNN's pre-sorted candidate pool for each user is used directly, with the top-ranked item serving as the supervision target.

\begin{figure*}[!ht]
    \centering
    \includegraphics[width=0.7\textwidth]{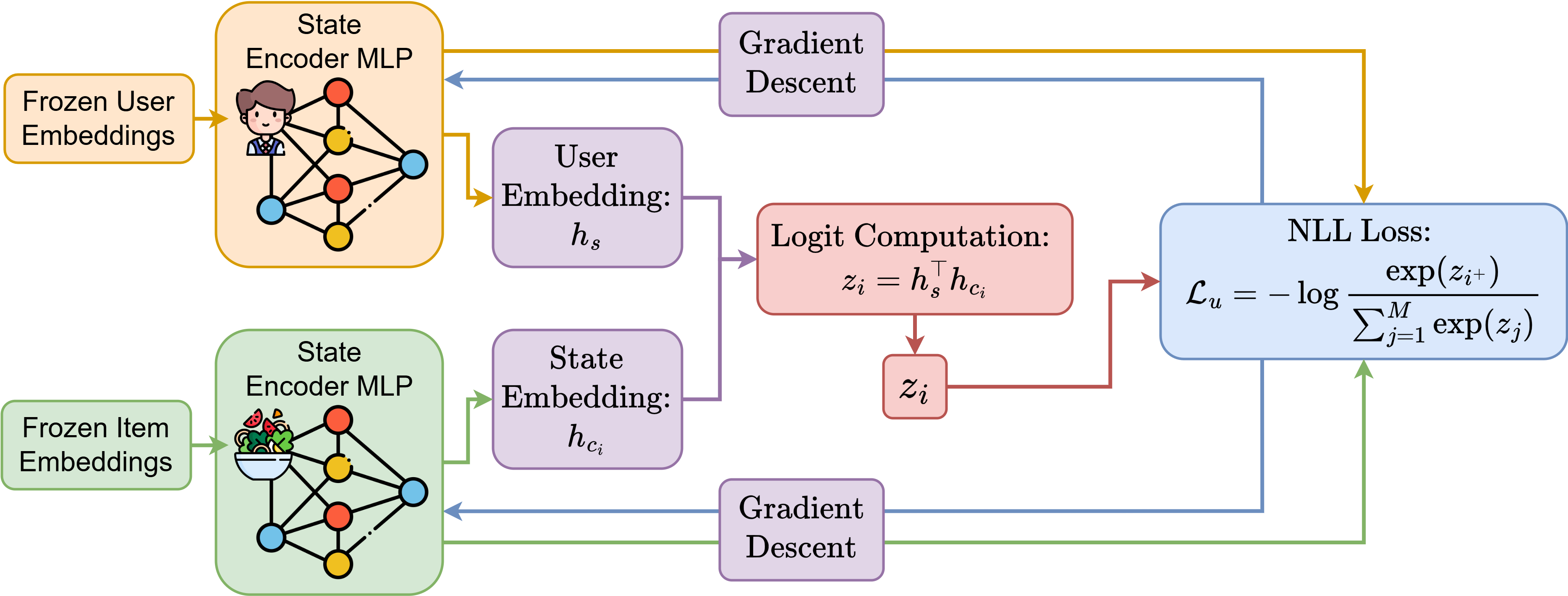}
    \caption{Behavior cloning pretraining schematic. Frozen user and item embeddings from MOPI-HFRS are passed through the state-encoder and candidate-encoder MLPs to produce $h_s$ and $h_{c_i}$. Logits $z_i = h_s^\top h_{c_i}$ feed a per-user NLL loss against the GNN's top-ranked item in the pool, and gradients update only the encoder MLPs; the underlying embeddings remain frozen.}
    \Description{A schematic of the behavior cloning pretraining stage, showing frozen user and item embeddings passed through state-encoder and candidate-encoder MLPs to produce encoded vectors, whose inner products form logits feeding a per-user negative log-likelihood loss against the GNN's top-ranked item.}
    \label{fig:bc-pretraining}
\end{figure*}

The BC stage uses the same actor architecture used during A2C: a state-encoder MLP applied to the (frozen) user embedding and a candidate-encoder MLP applied to the (frozen) candidate item embeddings. For each user, the candidate set is the GNN's pre-sorted pool of $M$ items. The supervision target is always the pool's top-ranked item (index 0), i.e., the item the GNN scores highest. Each user $u$ and candidate $c_i$ is encoded into
\[
h_s = \mathrm{MLP}_{\mathrm{state}}(h_u), \qquad
h_{c_i} = \mathrm{MLP}_{\mathrm{cand}}(e_{c_i}),
\]
and per-candidate logits are computed as inner products in the encoded space:
\[
z_i = h_s^\top h_{c_i}.
\]
The pretraining objective is the per-user negative log-likelihood of the GNN's top-ranked item under the softmax over the full pool:
\[
\mathcal{L}_u(\theta) = -\log \frac{\exp(z_{i^+})}{\sum_{j=1}^{M} \exp(z_j)}, \qquad
\mathcal{L}_{\mathrm{BC}}(\theta) = \mathbb{E}_u\!\left[\mathcal{L}_u(\theta)\right].
\]
This is a standard sampled-softmax / contrastive formulation: the policy must assign the highest score to whichever item the GNN ranked first, effectively distilling the GNN's ranking signal into the policy's scoring function. Figure~\ref{fig:bc-pretraining} summarizes the data flow during this stage.

Several aspects of this design are worth noting. First, the loss is computed against a single positive index per user rather than against a soft distribution over all items: the supervision signal is ``this is the correct choice'' rather than ``match this score distribution.'' This proves substantially more stable in practice than full-catalog distillation, both because it is computationally tractable and because it avoids forcing the policy to imitate the reference ranker's calibration on items that the user never actually interacted with. Second, only the actor's encoder MLPs are updated during BC; the underlying user and item embeddings remain frozen, consistent with the broader design principle of decoupling representation learning from policy learning. Third, the architecture used here is identical to the one used during A2C, so the pretrained weights transfer directly into the RL stage with no head-swap or projection layer required.

In our experiments, BC pretraining is run for a small number of epochs prior to A2C and produces a policy whose top-$K$ rollouts already approximate the reference ranker's top-$K$ recommendations. From this initialization, A2C refinement under the scalarized relevance/health reward perturbs the policy in directions that improve cumulative health alignment while preserving the bulk of its relevance behavior.

\subsection{Evaluation Procedure}

At evaluation time, the policy is run greedily for $K = 20$ steps per user, producing a top-$20$ list against which the same metrics used by MOPI-HFRS are computed (Recall@20, NDCG@20, H-Score@20). Training-time positive interactions are masked from the candidate pool, mirroring the masking convention in the original framework.

\paragraph{Use of AI Tools.}
Portions of the MORL-A2C implementation, including the reinforcement learning environment, actor-critic training loop, and evaluation harness, were developed with assistance of AI tools. All AI-generated code was reviewed, tested, and validated by the authors, who take full responsibility for the implementation and results.

\section{Experiments}

This section presents the empirical evaluation of MORL-A2C against the original MOPI-HFRS framework. We focus on the health-preference trade-off, evaluating three metrics: Recall@20 (relevance retrieval), NDCG@20 (rank-aware relevance), and H-Score@20 (personalized health alignment). We additionally describe the discovery and correction of an evaluation bug in the MOPI-HFRS reference implementation, which materially affects the comparison.

\subsection{Evaluation Protocol}

\begin{table}[t]
\centering
\caption{Hyperparameters for all reported experiments. Where the two benchmarks differ, values are shown as macro\,/\,all.}
\label{tab:hyperparams}
\setlength{\tabcolsep}{6pt}
\begin{tabular}{@{}ll@{}}
\toprule
\textbf{Parameter} & \textbf{Value} \\
\midrule
\multicolumn{2}{@{}l}{\textit{Behavior Cloning}} \\[2pt]
BC epochs              & 50\,/\,15 \\
BC learning rate       & $1 \times 10^{-3}$ \\
\midrule
\multicolumn{2}{@{}l}{\textit{A2C Training}} \\[2pt]
A2C epochs             & 5{,}000 \\
Learning rate          & $1 \times 10^{-4}$ \\
Batch size (users/epoch) & 64 \\
Health weight $\beta$  & 0.5 \\
Entropy weight $\lambda_e$   & 1.5\,/\,3.0 \\

Candidate pool size $M$ & full catalog \\
\midrule
\multicolumn{2}{@{}l}{\textit{Architecture}} \\[2pt]
State MLP hidden dim   & 256 \\
State MLP layers       & 2 \\
Candidate MLP layers   & 1 \\
Embedding dim $d$      & 128 \\
\bottomrule
\end{tabular}
\end{table}

Our evaluation pipeline mirrors the MOPI-HFRS protocol so that the comparison is faithful:

\begin{itemize}
    \item \textbf{Embedding computation.} A forward pass of the (frozen) GNN encoder produces user and food embeddings.
    \item \textbf{Top-$K$ generation.} For MOPI-HFRS, items are ranked by inner product between user and food embeddings, with training-time positive interactions masked. For MORL-A2C, the policy is rolled out greedily for $K = 20$ steps.
    \item \textbf{Metric computation.} Recall@20, NDCG@20, and H-Score@20 are computed against held-out positives and user health-tag profiles.
\end{itemize}

We use the same data splits, graph encoder, and tag scheme as MOPI-HFRS, isolating the change in how the top-$K$ list is constructed. Table~\ref{tab:hyperparams} lists the hyperparameters used for both benchmarks. We focus on three metrics whose baseline values we were able to faithfully reproduce from~\cite{MOPI_HFRS}. Two additional diversity-focused metrics reported in~\cite{MOPI_HFRS} (AvgTags@20 and \%Foods@20) exhibited substantial discrepancies between our reproduction and the published values (e.g., AvgTags@20 differed by approximately $2\times$), suggesting differences in metric computation or tag-matrix construction that we were unable to resolve. As our contribution targets the health-preference trade-off, we report the three validated metrics and discuss diversity-aware extensions in Section~7.

\subsection{Discovery and Correction of an Evaluation Bug in MOPI-HFRS}

In the course of integrating our RL evaluation harness with the MOPI-HFRS reference codebase, we identified an inconsistency in the evaluation pipeline that systematically understates baseline ranking performance. Specifically, two issues were observed:

\begin{enumerate}
    \item The GNN evaluation constructed a score matrix of shape $(\text{num\_test\_edges} \times |\mathcal{F}|)$, one row per test edge, not per unique user, because user embeddings were indexed by the negative-sampling source indices before the score matrix multiply. Retrieving recommendations via \path{top_K_items[user_id]} then used the actual user ID (e.g.\ 42) as a row index into this edge-indexed matrix, returning the recommendations for the 42\textsuperscript{nd} test edge rather than those for user 42.
    \item The exclusion mask used to remove already-observed interactions during top-$K$ ranking was constructed from \emph{negative-edge} index tensors used by the BPR sampler, rather than from the actual training-positive interaction set.
\end{enumerate}

When both issues are corrected (using the full embedding matrices for evaluation scoring and excluding genuine training-positive interactions from the candidate pool), reported ranking metrics for the underlying MOPI-HFRS model improve substantially. We refer to this as the ``MOPI-HFRS (bug-fixed)'' configuration in Table~\ref{tab:final-results}. Critically, this correction alters \emph{evaluation only}; no change is made to the MOPI-HFRS training procedure. We report results against this corrected baseline because comparing MORL-A2C to the artificially-low original numbers would overstate our method's gains.

\subsection{Results}

\begin{table}[t]
\centering
\caption{Comparison of MOPI-HFRS (original and bug-fixed evaluation) and MORL-A2C on both benchmarks. All values are percentages. Best Recall/NDCG is \underline{underlined}; best H-Score is \textbf{bolded}.}
\label{tab:final-results}
\setlength{\tabcolsep}{4pt}
\begin{tabular}{@{}lccc@{}}
\toprule
\textbf{Model} & \textbf{Recall@20} & \textbf{NDCG@20} & \textbf{H-Score@20} \\
\midrule
\multicolumn{4}{@{}l}{\textit{Nutrition-macro only}} \\[2pt]
MOPI-HFRS (as reported)    & 12.73 & 10.25 & 39.40 \\
MOPI-HFRS (bug-fixed)      & \underline{25.64} & \underline{23.52} & 46.05 \\
MORL-A2C                   & 23.61 & 20.64 & \textbf{69.57} \\
\midrule
\multicolumn{4}{@{}l}{\textit{Nutrition-all}} \\[2pt]
MOPI-HFRS (as reported)    & 12.92 & 10.14 & 61.13 \\
MOPI-HFRS (bug-fixed)      & \underline{25.34} & \underline{22.40} & 64.85 \\
MORL-A2C                   & 21.53 & 20.40 & \textbf{90.48} \\
\bottomrule
\end{tabular}
\end{table}

Table~\ref{tab:final-results} summarizes results on both benchmarks. Three observations stand out.

First, the bug-fixed MOPI-HFRS configuration substantially outperforms the originally-reported numbers across all three metrics on both benchmarks. On the macro benchmark, Recall@20 nearly doubles ($12.73 \rightarrow 25.64$) and NDCG@20 more than doubles ($10.25 \rightarrow 23.52$); similar gains appear on the full-nutrient benchmark. This reinforces the importance of treating evaluation infrastructure as a first-class research artifact: a non-trivial fraction of the reported gap between baseline and proposed methods in this domain may reflect harness-level discrepancies rather than methodological differences.

Second, MORL-A2C produces a modest reduction in ranking metrics relative to the corrected baseline.\footnote{We swept $\beta \in \{0.1, 0.3, 0.5, 1, 3, 5, 10, 20\}$ on the macro-nutrient validation set. $\beta = 0.5$ yielded the best balance of Recall@20 and H-Score@20; higher values further improved H-Score at the cost of sharper ranking degradation.} On the macro benchmark, Recall@20 drops by roughly two percentage points ($25.64 \rightarrow 23.61$) and NDCG@20 by approximately three points ($23.52 \rightarrow 20.64$). On the full-nutrient benchmark the pattern holds, with Recall@20 dropping by roughly four points ($25.34 \rightarrow 21.53$) and NDCG@20 by two points ($22.40 \rightarrow 20.40$). This is an expected consequence of the scalarized reward: the policy optimizes a combination of relevance and health alignment, trading some ranking quality for health gains by design.

Third, in exchange for that modest ranking concession, MORL-A2C achieves a substantial gain in personalized health alignment on both benchmarks. On the macro benchmark, H-Score@20 increases from $46.05$ to $69.57$ (an absolute improvement of over $23$ percentage points). On the full-nutrient benchmark the gain is even larger: H-Score@20 rises from $64.85$ to $90.48$ (an absolute improvement of over $25$ percentage points). This demonstrates that a sequential, policy-driven approach can effectively navigate the health-preference trade-off, buying meaningful improvement on a non-differentiable health metric at modest cost to ranking quality.

We note that a simpler non-learned reranking heuristic, for example greedily selecting the health-tag-maximizing item at each position from the GNN's top-$M$ candidates, could also improve H-Score@20 without RL. However, such a heuristic lacks the ability to learn position-dependent trade-offs from the reward signal; it would maximize health greedily at every step rather than balancing cumulative relevance and health across the full list. Evaluating such baselines to isolate the contribution of learned sequential optimization is a natural next step.

\section{Conclusion}

We presented MORL-A2C, a multi-objective reinforcement learning reranker that extends the MOPI-HFRS framework by augmenting its static, MGDA-based Pareto optimization with a sequential, policy-driven decision process targeting the health-preference trade-off. By formulating top-$K$ recommendation as a $K$-step MDP over frozen GNN embeddings, leveraging an Advantage Actor-Critic algorithm with a scalarized relevance/health reward, and warm-starting the policy via behavior cloning, MORL-A2C is able to optimize set-level objectives such as cumulative health-tag coverage that single-shot rankers cannot natively express. We additionally identified and corrected a non-trivial evaluation bug in the original MOPI-HFRS reference implementation, and reported all results against the corrected baseline.

Empirically, MORL-A2C delivers a modest reduction in ranking metrics ($\sim$2 points Recall@20, $\sim$3 points NDCG@20) in exchange for a substantial improvement in personalized health alignment (H-Score@20 from $46.05$ to $69.57$), with consistent trends observed on both the macro-nutrient and full-nutrient benchmarks.

These findings suggest that policy-space optimization is a productive direction for health-aware recommendation. By moving from one-shot static ranking to sequential decision-making, the system gains a vocabulary in which set-level health objectives can be expressed and optimized directly. The limitations discussed in Section~7---most notably the absence of an explicit diversity reward and the reliance on offline evaluation---define clear next steps; we view the sequential formulation itself, rather than any specific algorithmic choice, as the central contribution of this work.

\section{Limitations \& Future Work}

While MORL-A2C demonstrates that policy-driven sequential optimization can effectively navigate the health-preference trade-off, several limitations remain.

\begin{description}
\item[Diversity.]
The current reward includes no diversity signal, and preliminary inspection confirms that the policy concentrates recommendations on a narrow subset of the food catalog. Adding a third reward component penalizing pairwise embedding similarity within an episode, mirroring $\mathcal{L}_{\mathrm{diversity}}$ in~\cite{MOPI_HFRS}, is the most immediate extension.

\item[Metric Reproducibility.]
Two diversity-focused metrics from~\cite{MOPI_HFRS} (AvgTags@20 and \%Foods@20) exhibited substantial discrepancies between our reproduction and the published values. Resolving these is a prerequisite for validating any future diversity reward component.

\item[Dynamic Scalarization.]
The fixed $\beta$ universalizes the relevance--health trade-off across all users. Learning a state-conditioned $\beta(s_t)$~\cite{DRS} would allow per-user and per-position adaptation.

\item[Online Feedback.]
Rewards currently come from a simulated offline environment; the policy does not observe actual sequential user feedback. Extending MORL-A2C to an online or hybrid setting would more faithfully capture real dietary decision-making.

\end{description}

\bibliographystyle{ACM-Reference-Format}
\bibliography{references}

\end{document}